\documentclass[runningheads]{llncs}

\usepackage{eccv}

\usepackage{eccvabbrv}

\usepackage{graphicx}
\usepackage{booktabs}
\usepackage{makecell}
\usepackage[table]{xcolor}
\usepackage{minitoc}
\usepackage{titletoc}
\usepackage{algorithm}
\usepackage{algorithmicx}
\usepackage{algpseudocode}
\usepackage[pagebackref]{hyperref}
\usepackage[accsupp]{axessibility}  
\usepackage{xcolor}

\usepackage{hyperref}

\usepackage{orcidlink}

\begin{document}

\title{MultiWorld: Scalable Multi-Agent Multi-View Video World Models}
\titlerunning{MultiWorld}

\author{Haoyu Wu\inst{1} \and
Jiwen Yu\inst{1} \and
Yingtian Zou\inst{2} \and
Xihui Liu\inst{1}}

\authorrunning{H.~Wu et al.}

\institute{The University of Hong Kong \and
Sreal AI}

\maketitle

\begin{abstract}
Video world models have achieved remarkable success in simulating environmental dynamics in response to actions by users or agents. They are modeled as action-conditioned video generation models that take historical frames and current actions as input to predict future frames. Yet, most existing approaches are limited to single-agent scenarios and fail to capture the complex interactions inherent in real-world multi-agent systems. We present \textbf{MultiWorld}, a unified framework for multi-agent multi-view world modeling that enables accurate control of multiple agents while maintaining multi-view consistency. We introduce the Multi-Agent Condition Module to achieve precise multi-agent controllability, and the Global State Encoder to ensure coherent observations across different views. MultiWorld supports flexible scaling of agent and view counts, and synthesizes different views in parallel for high efficiency. Experiments on multi-player game environments and multi-robot manipulation tasks demonstrate that MultiWorld outperforms baselines in video fidelity, action-following ability, and multi-view consistency.


\end{abstract}

\begin{figure}[h]
    \centering
    \includegraphics[width=\linewidth, clip]{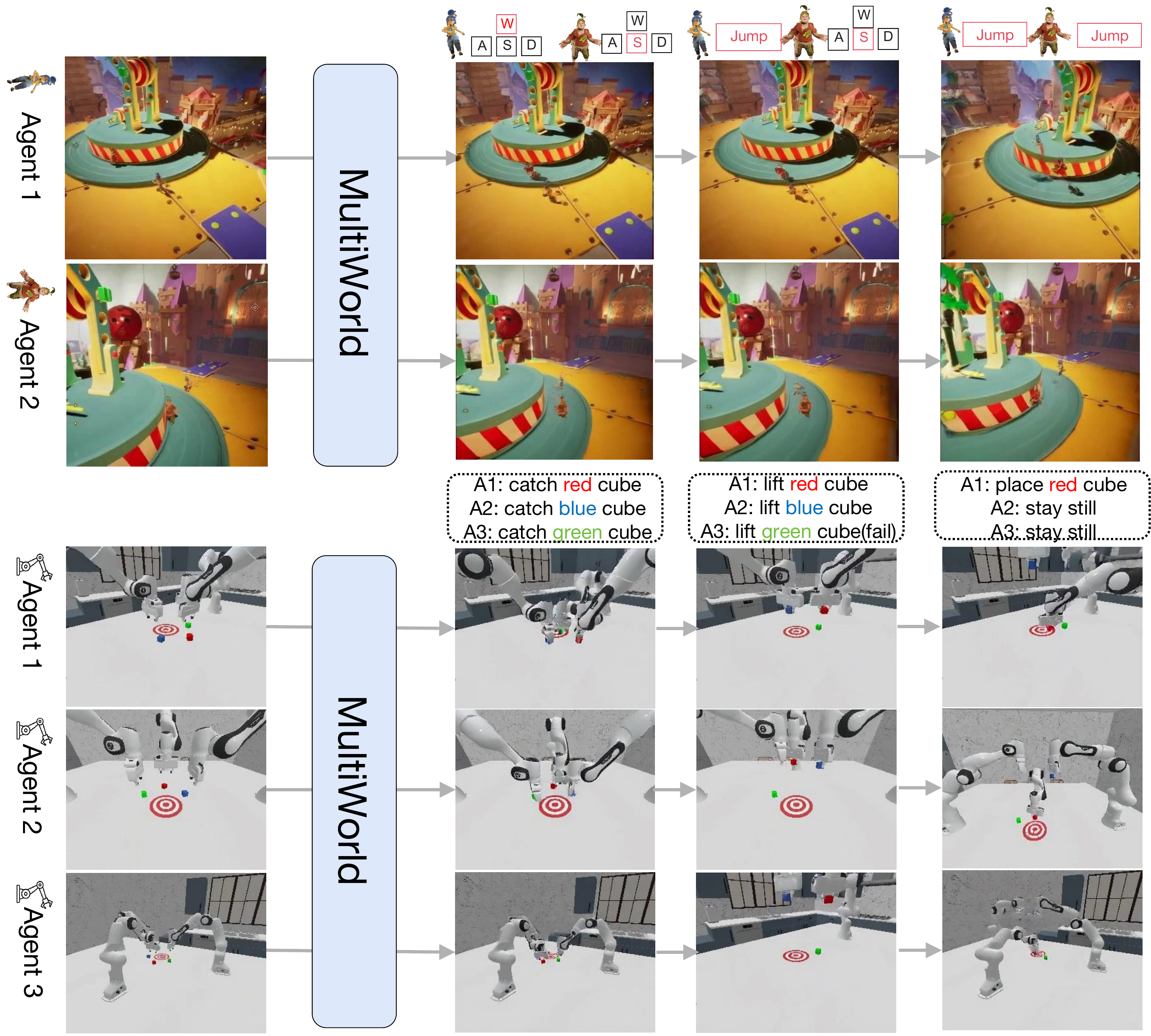}
    \caption{\textbf{MultiWorld generates multi-agent multi-view videos. } Given initial views and per-frame actions, our model produces action-controllable, multi-view consistent videos in both multi-player video game and multi-robot manipulation scenarios.}
    \label{fig:teaser}
\end{figure}

\section{Introduction}

Video world models~\cite{lingbo-worldmodel,kong2024hunyuanvideo,mao2025yume,huang2025vid2world,wang2024worlddreamer,yang2026stableworld,wu2025video} have achieved significant success in accurately predicting future environment dynamics conditioned on text or actions. However, existing video world models implicitly assume a single agent in the simulated environment, ignoring interactions and interdependencies among multiple agents acting simultaneously, as in collaborative robotics and multi-player video games. Furthermore, since each agent perceives the shared environment from its own viewpoint, multi-agent simulation inherently requires generating consistent observations across multiple views. However, previous models fail to preserve scene consistency among multiple views observed by different agents. 

Extending world models to multi-agent, multi-view scenarios introduces three key challenges that existing single-agent world models cannot address: (1) \textbf{Multi-Agent Controllability}: controlling multiple agents requires associating specific actions with corresponding agents and synchronizing their executions. (2) \textbf{Multi-View Consistency}: since multiple agents inherently have different partial observations of a shared environment from distinct viewpoints, the model must synthesize visually coherent videos across diverse perspectives, while ensuring observations from multiple agents remain geometrically consistent. (3) \textbf{Framework Scalability}: real-world environments involve variable numbers of agents and views, requiring a framework that generalizes across configurations without assuming fixed agent counts or camera setups. Previous works~\cite{zhang2024combo,solaris2026} assume a fixed number of agents or predefined camera views, which limits their applicability in diverse real-world scenarios.

In this work, we introduce MultiWorld to address the challenges of multi-agent, multi-view world modeling, enabling flexible scaling of agent and view counts, as illustrated in Fig.~\ref{fig:teaser}. (1) To achieve Multi-Agent Controllability, we propose the \textbf{Multi-Agent Condition Module (MACM)} (Sec.~\ref{sec:macm}), which consists of Agent Identity Embedding (AIE) and Adaptive Action Weighting. Simply stacking different agents' actions together makes it difficult for the video world model to associate input actions with corresponding agents. To address this, we propose AIE to overcome the limitations of agent-agnostic architectures by injecting distinct identity embeddings into action tokens. In addition, we introduce an Adaptive Action Weighting mechanism that dynamically prioritizes active agents over static ones, guiding the model to focus on dynamic actions and state changes. (2) For Multi-View Consistency, we aggregate different observations into a global 3D-aware environment state via the \textbf{Global State Encoder (GSE)} (Sec.~\ref{sec:gse}). This global state ensures each view synthesis remains anchored to a consistent environmental state, inherently enforcing coherence across perspectives. (3) Our framework (Sec.~\ref{sec:inference}) is scalable with respect to both agent count and view count. Scalability with respect to the number of agents is achieved by the Multi-Agent Conditioning Module, which assigns each agent a relative identity embedding and computes action weights independently for each agent, enabling the framework to be extended naturally to any number of agents without architectural changes. Scalability with respect to the number of views is achieved by the Global State Encoder, which encodes any number of multi-view observations into a 3D-aware global state condition for arbitrary view generation, decoupling computation from the number of views and enabling flexible integration of additional viewpoints. By encoding observations into a compact global state, GSE further enables efficient parallel and autoregressive simulation of arbitrary numbers of camera views.

We construct two multi-agent, multi-view datasets in multi-player games and collaborative robotics. First, we collect a real-player dataset from the game \textit{ItTakesTwo}, which features a dual-agent scenario with distinct viewpoints and interdependent actions. Beyond dual-agent settings, we collect several thousand episodes in a multi-robot manipulation simulator using RoboFactory~\cite{qin2025robofactory}, in which the number of agents and camera viewpoints vary. Together, these datasets cover both complex inter-agent interactions and variable agent-view configurations, enabling comprehensive evaluation of MultiWorld. Extensive experiments demonstrate that MultiWorld consistently outperforms competitive baselines in video quality, action controllability, and multi-view consistency.

In summary, our key contributions include:

\begin{itemize}

    \item We propose \textbf{MultiWorld}, a unified framework for scalable multi-agent, multi-view world models that enables precise action adherence and synchronized simulation in the presence of complex cross-view dependencies.
    
    \item We introduce the Multi-Agent Condition Module (MACM) and the Global State Encoder (GSE), which jointly address the challenges of agent controllability, multi-view consistency, and scalability with the number of agents and views.

    \item We construct two complementary multi-agent simulation datasets based on a multi-player video game and a scalable robotics simulator, providing comprehensive benchmarks for variable agent configurations. Extensive experiments demonstrate that MultiWorld consistently outperforms competitive baselines.

\end{itemize}

\section{Related Work}

\noindent\textbf{Interactive Video World Models.} Diffusion models~\cite{ho2020denoising, rombach2022high,peebles2023scalable,lipman2023flow,bruce2024genie,parker2024genie2,genie3,alonso2024atari} have positioned video generation as a promising approach to world modeling. Beyond text-to-video synthesis~\cite{genie3,lingbo-worldmodel,wan2025wan,liu2025videodpo,ye2025fast}, interactive video generation~\cite{IGVsurvey} that responds to interactive control signals has evolved rapidly. Existing models incorporate various signals like camera controls~\cite{he2024cameractrl,Yu2024ViewCrafterTV,dfot,wu2025geometryforcing} and action controls~\cite{oasis2024,guo2025mineworld,feng2024matrix,shin2024wham} to simulate future states. Recent studies have explored several essential properties~\cite{huang2025towards} of interactive video world models, such as physical consistency~\cite{wang2025wisa,wang2025enhancing,zheng2025world4drive0,song2025physical}, and long-horizon coherence~\cite{yu2025context,xiao2025worldmem,wu2026infinite}, alongside efficient real-time generation ~\cite{yin2023one0step,huang2025self,zhu2026causal,yuan2026helios0} to enable practical deployment. With these properties, world models can serve as powerful simulators for downstream tasks like game generation~\cite{tang2025hunyuan,xiang2026geometryrope}, embodied AI~\cite{nvidia2025world,chi2025wow0}, and autonomous driving~\cite{russell2025gaia,yang2025geniedrive0}. Game video world models~\cite{yu2025gamefactory,valevski2024diffusion} control the environment and simulate player observations based on provided actions. Robotic video world models serve as a more generalizable virtual environment~\cite{zhu2025irasim,zheng2025flare,lyu2025dywa,zhu2025wmpo0} to help develop better vision-language-action (VLA) models~\cite{zhen2025tesseract,jiang2025world4rl,lykov2025physicalagent,yang2024physical}. Despite promising progress across these domains, existing approaches primarily focus on single-agent settings, overlooking the fact that real-world environments typically involve multiple agents sharing a single environment. In this work, we build a multi-agent video world model that supports variable numbers of agents acting within a unified environment to narrow this critical gap.

\noindent\textbf{Multi-Agent Planning and Simulation.} Multi-agent planning~\cite{feng2025multiagentembodiedai} aims to coordinate multiple agents in a shared environment to achieve individual or shared goals. Multi-agent planning requires an environment for execution. RoboFactory~\cite{qin2025robofactory} investigated reasonable and safe multi-agent cooperation with compositional constraints in robotic manipulation tasks. SeqWM-MARL~\cite{zhao2025empowering} employs a sequential paradigm for multi-agent reinforcement learning in robotics tasks. Similarly, TeamCraft~\cite{long2024teamcraft} and CausalMACE~\cite{chai2025causalmace} developed systems for multi-agent cooperation within Minecraft environments.

However, these works rely on physical simulators or game engines that require complex manual environment design. In contrast, video world models offer a generalizable alternative for simulating multi-agent environments, yet multi-agent video world models remain underexplored. Developing a multi-agent simulator presents several challenges, including controllability among multiple agents and ensuring multi-view consistency. MultiVerse~\cite{enigma2025multiverse} simulates a two-player racing game. COMBO~\cite{zhang2024combo} attempts to integrate several single-agent video world models into a single multi-agent model, but this approach fundamentally neglects inter-agent interactions.

Our concurrent work~\cite{solaris2026} builds a two-player video world model in Minecraft. Their dataset assumes a fixed number of agents and views, precluding variable agent configurations. Their model interleaves observations from two views along the sequence dimension and processes them through shared self-attention, but this approach cannot scale to more views due to computational and memory constraints. In contrast, MultiWorld supports scalable multi-agent and multi-view video world modeling through our Global State Encoder, which compresses cross-view information into a compact latent representation.

\begin{figure}[ht]
    \centering
    \includegraphics[
        width=\linewidth,
        clip
    ]{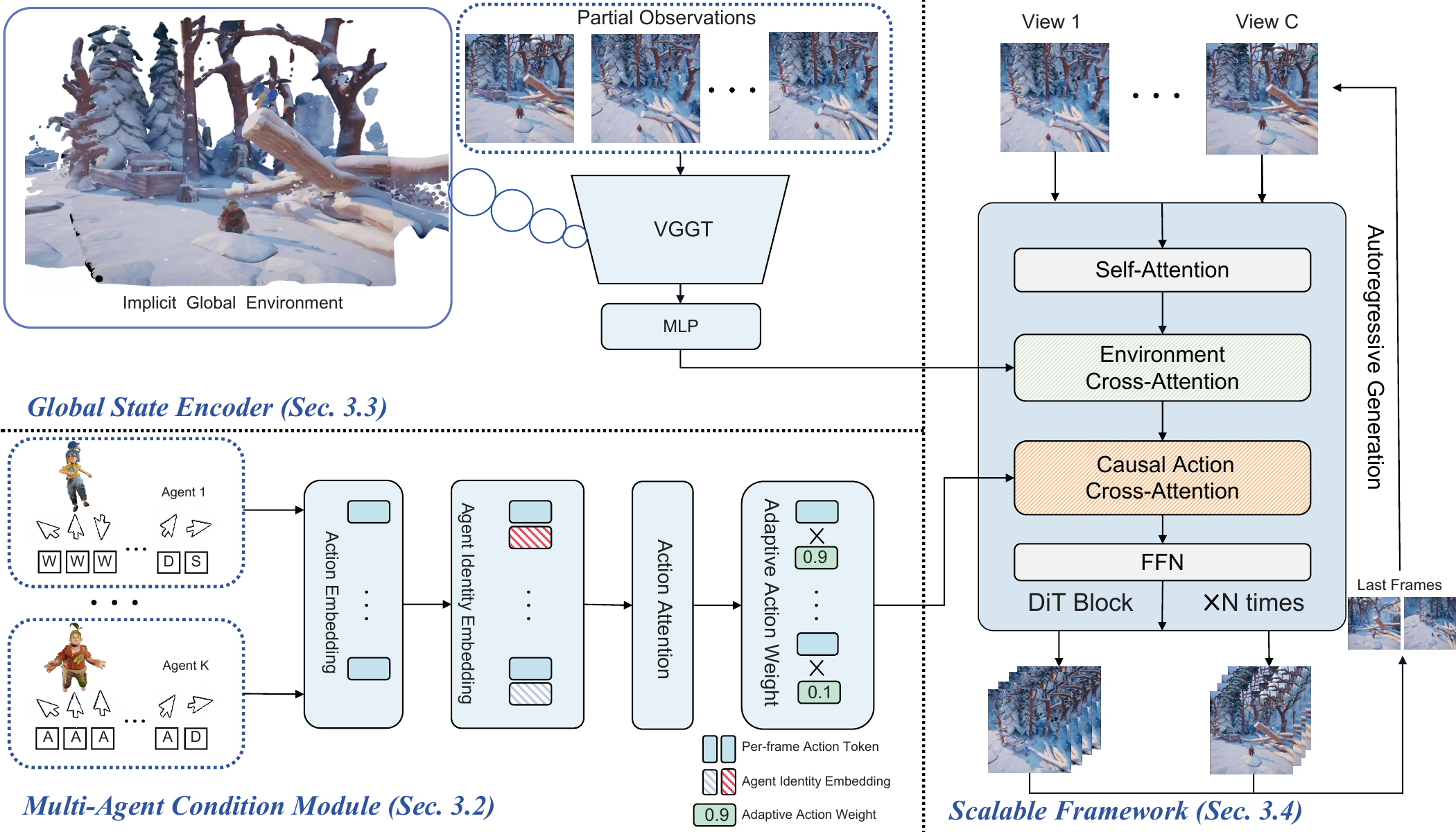}
    \caption{\textbf{Pipeline of MultiWorld.} We propose MultiWorld, a unified framework for scalable multi-agent multi-view video world modeling. In the Multi-Agent Condition Module (Sec.~\ref{sec:macm}), Agent Identity Embedding and Adaptive Action Weighting are employed to achieve multi-agent controllability. In the Global State Encoder (Sec.~\ref{sec:gse}), we use a frozen VGGT backbone to extract implicit 3D global environmental information from partial observations, thereby improving multi-view consistency. MultiWorld scales effectively across varying agent counts and camera views, supporting autoregressive inference to generate beyond the training context length (Sec.~\ref{sec:inference}).}
    \label{fig:pipeline}
\end{figure}

\section{Method}
\label{sec:method}
This section presents MultiWorld, a multi-agent, multi-view video world model that generates multi-view-consistent videos conditioned on multi-agent actions. The framework consists of an action-conditioned diffusion backbone (Sec.~\ref{sec:acvdm}), the Multi-Agent Condition Module (MACM) for scalable multi-agent control (Sec.~\ref{sec:macm}), and the Global State Encoder (GSE) for multi-view consistency (Sec.~\ref{sec:gse}), as illustrated in Fig.~\ref{fig:pipeline}. We further discuss the scalability of the pipeline in terms of agent and view counts in Sec.~\ref{sec:inference} and describe our autoregressive strategy for parallel and long-horizon world simulation.

\subsection{Backbone and Notation}
\label{sec:acvdm}

We formulate multi-agent, multi-view world simulation as a collection of $C$ image-action-conditioned video generation problems, where $C$ is the number of camera views. Videos from different camera views are synthesized in parallel with a shared global environment state. Our model is built on Flow Matching (FM)~\cite{lipman2023flow, liu2023flow} with a Transformer backbone~\cite{vaswani2017attention}. We consider $K$ agents and $C$ camera views, where $K$ and $C$ are independent. Let $a_i = (a_i^1,\dots,a_i^K)$ denote the joint action of all $K$ agents at frame $i$, and let $\mathbf{a} = \{a_0,\dots,a_I\}$ denote the full action sequence. The video recorded by camera $c \in \{1,\ldots,C\}$ is denoted as $\mathbf{x}_{c}$. We define the \emph{environment observation} as $\mathbf{o} = \{\mathbf{o}_{c}\}_{c=1}^{C}$, where $\mathbf{o}_{c}$ is the initial frame of $\mathbf{x}_{c}$; these initial frames supply global scene context for multi-view world modeling and are updated as the environment evolves.

For each camera view $c$, we apply FM to model the conditional distribution of future video frames. Sampling $t \sim \mathcal{U}(0,1)$ and noise $\boldsymbol{\epsilon} \sim \mathcal{N}(\mathbf{0},\mathbf{I})$, the noisy observation and target velocity are defined as:
\[
  \mathbf{x}_{c}^{t} = (1-t)\,\mathbf{x}_c + t\,\boldsymbol{\epsilon},
  \qquad
  \mathbf{u} = \boldsymbol{\epsilon} - \mathbf{x}_c.
\]
We parameterize the velocity network as $v_\theta(\mathbf{x}_{c}^{t},\,t,\,\mathbf{a},\,\mathbf{o})$, trained to predict the target velocity $\mathbf{u}$ conditioned on the noisy observation, timestep $t$, action sequence $\mathbf{a}$, and environment observation $\mathbf{o}$.

To enforce temporal causality, we apply a frame-wise causal mask to the action cross-attention, as illustrated in Fig.~\ref{fig:pipeline}. This ensures that video tokens at frame $i$ attend only to actions from frames $\{0,\ldots, i\}$, preventing future information leakage and supporting stable long-horizon autoregressive generation.

\subsection{Multi-Agent Condition Module}
\label{sec:macm}

\noindent\textbf{Observation.} A multi-agent video world model must distinguish different agent actions and simulate interactions between agents. However, two core challenges arise. First, simply stacking multiple identical action encoders leads to identity ambiguity. For instance, the model may struggle to process ``mirror actions'' (e.g., Agent 1 moving left while Agent 2 moving right, versus the reverse). Second, as agents' action strengths vary over time (some agents are actively moving while others remain stationary), treating all agent actions equally may cause models to ignore the primary drivers of environmental change.

The Multi-Agent Condition Module first embeds the actions into latent action tokens and assigns Agent Identity Embedding (AIE) to address identity ambiguity. Self-attention is applied to the agent action tokens to explicitly model interactions among agents. The action tokens are subsequently aggregated with weights into a unified action token for each frame with Adaptive Action Weighting (AAW). Finally, these aggregated action tokens are injected into the DiT backbone through causal cross-attention. 

\noindent\textbf{Agent Identity Embedding.}
\label{paragraph:AIE}
To resolve identity ambiguity, we introduce Agent Identity Embedding (AIE). To support a variable number of agents, we adopt Rotary Position Embedding~\cite{rope} to compute agent identity embeddings. 

Specifically, given an action latent at frame $f$, $\mathbf{a_f} \in \mathbb{R}^{K \times D}$, where $D$ denotes the latent dimension, and we omit the frame subscript since AIE is independent of the frame index. For each agent $i \in \{1, \dots, K\}$, we apply a rotation matrix $\mathbf{R}_{\Theta, i}$ to the action embedding $a_f^i \in \mathbb{R}^D$:

\begin{equation}
    \text{AIE}(a_i, i) = \mathbf{R}_{\Theta, i} a_i
\end{equation}

where $\mathbf{R}_{\Theta, i}$ is defined by pre-computed frequencies $\theta_j = b^{-2j/D}$, and $b$ is the base frequency constant following the standard RoPE formulation. Specifically, for each pair of dimensions $(2j, 2j+1)$, the transformation is:

\begin{equation}
\begin{pmatrix} 
a^{(2j)} \\ a^{(2j+1)} 
\end{pmatrix}_{out} = 
\begin{pmatrix} 
\cos(i\theta_j) & -\sin(i\theta_j) \\ 
\sin(i\theta_j) & \cos(i\theta_j) 
\end{pmatrix}
\begin{pmatrix} 
a^{(2j)} \\ a^{(2j+1)} 
\end{pmatrix}_{in}
\end{equation}

By applying AIE along the agent dimension, we break the symmetry of the multi-agent action space. We then adopt a self-attention block to model inter-agent interactions. The attention between agents $m$ and $n$ is computed as:
\begin{equation}
(\mathbf{R}_m \mathbf{a_m})^\top (\mathbf{R}_n \mathbf{a_n}) = \mathbf{a_m}^\top \mathbf{R}_m^\top \mathbf{R}_n \mathbf{a_n} = \mathbf{a_m}^\top \mathbf{R}_{n-m} \mathbf{a_n}. 
\end{equation}

Thus, AIE effectively eliminates identity confusion in video world modeling.

\noindent\textbf{Adaptive Action Weighting.} The influence of different agents varies: at any given moment, some agents may be stationary while others are actively moving. We propose an Adaptive Action Weighting to prioritize active agents over static ones. We implement this via a multi-layer perceptron (MLP) that predicts adaptive weighting factors for each action token. The action tokens are multiplied by their corresponding weighting factors and then summed to form a unified representation. By dynamically scaling each agent's action embedding, the model can more effectively focus on the dynamic agents that drive the environmental change more significantly.

\subsection{Global State Encoder}
\label{sec:gse}

\noindent\textbf{Observation.} The multi-agent multi-view video world model functions as a shared environment that receives all agent actions and renders corresponding partial observations from multiple distinct views. Therefore, it is crucial to preserve spatial consistency among observations from different viewpoints.

To achieve view consistency, the video world model is expected to maintain the global environmental information and to predict new frames conditioned on such information. Moreover, we need 3D-aware global environmental information to ensure 3D consistency. Motivated by this, we employ a pretrained VGGT~\cite{wang2025vggt}, an end-to-end 3D reconstruction foundation model, as the backbone of our proposed \emph{Global State Encoder} (GSE) for extracting 3D-aware environmental states. Given multiple input views, VGGT produces latent representations of the current global environment state that can be reconstructed into a 3D scene, and these latent representations are fused through a multi-layer perceptron (MLP) into a compact global representation. 

Formally, given a multi-view observation set $\mathbf{O} = \{\mathbf{O}_c\}_{c=1}^C$ comprising $C$ images $\mathbf{O}_c \in \mathbb{R}^{3 \times H \times W}$ that capture the same environment at any given time, the GSE first encodes these variable-length observations using the VGGT backbone. This process yields latent features $\mathbf{H}_{\text{vggt}} = \text{VGGT}(\mathbf{O})$, where $\mathbf{H}_{\text{vggt}} \in \mathbb{R}^{C \times n \times d}$, $n$ denotes the number of tokens per image, and $d$ represents the latent dimension. These features are subsequently processed by an MLP to align their dimensions with the DiT backbone, resulting in $\mathbf{H} = \text{MLP}(\mathbf{H}_{\text{vggt}})$. Finally, $\mathbf{H}$ is injected into the DiT via cross-attention to condition the generation process, enabling the diffusion model to generate multi-view-consistent partial observations based on the shared global environment state. It should be noted that we do not explicitly reconstruct a 3D point cloud as the global environment state, but leverage the latents $\mathbf{H}_{\text{vggt}}$ inherently contain the 3D spatial information.

This design offers three advantages: (1) improving multi-view spatial consistency by a shared global representation; (2) supporting an arbitrary number of views by compressing multi-view information into a unified 3D-aware global environment state; and (3) supporting efficient prediction of different views by parallel generation.

\subsection{Scalable Framework}
\label{sec:inference}
Our MultiWorld framework is scalable to arbitrary numbers of agents and camera views.

The agent scalability is achieved by AIE described in Sec.~\ref{sec:macm}. AIE assigns a relative identity embedding to each agent to model inter-agent relationships. Because these relative embeddings can be effectively extrapolated, the framework is able to accommodate an arbitrary number of agents without performance degradation. 

MultiWorld decomposes multi-view simulation into a collection of single-view image-action-conditioned video generation tasks with a shared global environmental state. The framework achieves camera-view scalability by compressing a variable number of partial observations into a global environment state with the GSE (Sec.~\ref{sec:gse}), ensuring robust performance regardless of the input view count. 

Besides, decomposing multi-view simulation into a series of single-view simulation conditions on the global context allows videos from different camera views to be generated simultaneously. This leads to another advantage of our framework: inference latency remains nearly constant regardless of the view count when computational resources are scaled accordingly. The framework for parallelized view generation achieves an approximate $1.5\times$ speedup over sequential generation in double-view simulation. 

Beyond camera-view scalability, the framework supports stable, long-horizon multi-view simulation via autoregressive generation. We autoregressively simulate multi-view video chunks and update the global environment state regularly. By iteratively feeding the most recent observations back into the Global State Encoder (GSE), the global environment state is continuously updated. In practice, we generate the first chunk of all camera views and use the last frames to update the global environment state for the next chunk generation. This approach enables stable generation for temporal horizons that exceed twice the training context length, with minimal performance degradation (see Fig.~\ref{fig:longvideo}). 

\section{Experiment}
In this section, we evaluate the MultiWorld across two distinct domains: multi-player video games and multi-robot manipulation. We conduct a fair comparison between MultiWorld and baseline methods in a dual-player video game. Then we extend MultiWorld to a multi-robot scenario to validate its scalability with respect to the number of agents and cameras. We demonstrate that MultiWorld generalizes effectively across domains and shows strong potential for a broad range of downstream applications.

\paragraph{Dataset and Metrics.} In our video game dataset, we record 500 hours of real-player data from \textit{ItTakesTwo} at 60 fps. After preprocessing, we retain only 100 hours with clear actions and stable camera motion. This effort resulted in a large-scale multi-player dataset characterized by high visual quality, containing over 21 million frames at an original resolution of 2560 x 1440. For the robotic dataset, we constructed a multi-robot manipulation dataset using \textit{RoboFactory}, which encompasses several tasks involving 2 to 4 agents. Comprehensive details on dataset construction are provided in the supplementary material (Sec.~\ref{supp:dataset}). We evaluate visual quality through metrics such as FVD (Fréchet Video Distance)~\cite{unterthiner2018towards}, PSNR (Peak Signal-to-Noise Ratio), SSIM (Structural Similarity Index)~\cite{wang2004image}, and LPIPS (Learned Perceptual Image Patch Similarity)~\cite{zhang2018unreasonable}. To assess multi-view consistency, we compute the reprojection error (RPE)~\cite{wu2025geometryforcing} across synchronized viewpoints. Furthermore, we evaluate action-following ability using the Inverse Dynamics Model (IDM) following VPT~\cite{baker2022vpt}.

\subsection{Baselines}

We compare MultiWorld against three representative baselines: (1) \textbf{Standard Image-Action-to-Video Model (Standard)}: This approach trains a video generation model treating each view independently as the most direct extension of a single-agent world model to a multi-agent and multi-view world model. (2) \textbf{Concatenated-View Video World Model (Concat-View)}: This approach combines different views into a single video for datasets with a fixed number of views, leveraging pretrained VAEs to provide basic cross-view information. However, it has significant drawbacks, including infeasibility in training due to GPU memory constraints as the number of views increases, and an inability to handle a variable number of views. (3) \textbf{COMBO}~\cite{zhang2018unreasonable}: This approach train compositional multi-agent world model with two stages: initially training several single-agent models, followed by the composition of these models into a multi-agent framework. While COMBO treats each agent independently and combines their dynamics, it does not address inter-agent interactions, which can lead to suboptimal performance in action-following tasks.

\subsection{Implementation Details}
We train MultiWorld using the Wan2.2-5B model~\cite{wan2025wan} on 81 frames. The resolution is set to 320 x 320 for each view in the video game and 320 x 256 for the robotics dataset. The training process consists of 40,000 iterations with a learning rate of 5e-5, employing a cosine learning rate scheduler and a global batch size of 64 across 8 NVIDIA A800 GPUs. The entire training takes approximately 4 days. More details can be found in the Supp.~\ref{supp:training}.

\subsection{Main Experiments}
We compare our model with state-of-the-art approaches on both the multi-player video game and multi-robot manipulation datasets. The evaluation results demonstrate the effectiveness and generalization ability of our MultiWorld framework in both video game and robotic manipulation scenarios. We report results in Tab.~\ref{tab:multi_view_comparison}. The improvements also show that the modules described in Sec.~\ref{sec:method} cooperate well and yield better results. We provide further ablations to decompose the contribution of each component.

As shown in Tab.~\ref{tab:multi_view_comparison}, our method consistently outperforms all baselines on FVD, PSNR, action-following, and RPE metrics in both multi-player video game and multi-robot manipulation scenarios. The results demonstrate the effectiveness of the MultiWorld framework and highlight its outstanding performance.

\begin{table}[htbp]
\centering
\setlength{\tabcolsep}{7pt}
\renewcommand{\arraystretch}{1.3}  
\caption{Comparison of different multi-view conditioning strategies across two scenarios: robot manipulation and multi-player video game. Our method (MultiWorld) achieves the best performance across most metrics. \textbf{bold} values denote the best, and \underline{Underlined} values indicate the second best. *indicates the method is not comparable with others because only trained on two camera views per episode. }
\begin{tabular}{@{}lcccccc@{}}
\toprule
\textbf{Method} & \textbf{FVD$\downarrow$} & \textbf{LPIPS$\downarrow$} & \textbf{SSIM$\uparrow$} & \textbf{PSNR$\uparrow$} & \textbf{Action$\uparrow$} & \textbf{RPE$\downarrow$} \\
\midrule
\rowcolor{gray!15} \multicolumn{7}{l}{Multi-Player Video Game} \\
\midrule
Standard & 245 & 0.36 & \underline{0.50} & 17.48 & 88.4 & 0.75 \\
Concat-View & 215 & 0.36 & 0.49 & 17.54 & 89.1 & 0.74  \\
Combo & 207 & \textbf{0.34} & \textbf{0.51} & \textbf{17.82} & \underline{89.3} & \underline{0.72} \\
Ours & \textbf{179} & \underline{0.35} & \textbf{0.51} & \underline{17.72} & \textbf{89.8} & \textbf{0.67 }\\
\midrule
\rowcolor{gray!15} \multicolumn{7}{l}{Multi-Robot Manipulation} \\
\midrule
Standard & 100  & \textbf{0.07} & \textbf{0.90} &26.39 & 88.2 & 1.60 \\
Concat-View* & 106 & 0.06 & 0.90 & 27.44 & 92.0 & 0.82  \\
Combo & \underline{99} & \underline{0.08} & \textbf{0.90} & \underline{26.49 }& \underline{88.5} & 1.54  \\
Ours & \textbf{96} & \textbf{0.07} & \textbf{0.90} & \textbf{26.60} & \textbf{88.7} & \textbf{1.52} \\
\bottomrule
\end{tabular}
\label{tab:multi_view_comparison}
\end{table}

\subsection{Ablation Study}
\label{sec:ablation}
We conduct comprehensive ablation experiments to decompose the effectiveness of each component of MultiWorld, as described in sec~\ref{sec:method}. The analysis proceeds from architectural necessities to the design choices for each component, systematically isolating the contribution of each proposed module.

\paragraph{Main Architectural Components.} We ablate the Multi-Agent Condition Module (MACM) and Global State Encoder (GSE) by comparing three variants: (1) a standard image-action-to-video baseline, (2) $+$ MACM, and (3) Both MACM and GSE modules applied. We evaluate each on visual quality and action-following ability to assess improvements in motion fidelity, perceptual quality, and global coherence. The results show that introducing MACM significantly improves action-following ability, and introducing GSE improves multi-view consistency, with both modules contributing to improved visual quality. 

\begin{table}[t]
\centering
\small
\setlength{\tabcolsep}{8pt}
\caption{\textbf{Ablation Study of Main Architectural Components.} We conduct the ablation by adding MACM and GSE on the standard image-action-to-video baseline step-by-step. The results indicate that incorporating MACM enhances action controllability. Integrating GSE improves multi-view consistency. Improvement in action controllability and multi-view consistency leads to improved visual quality.}

\begin{tabular}{lcccccc}
\toprule
Config & FVD $\downarrow$ & LPIPS $\downarrow$ & SSIM $\uparrow$ & PSNR $\uparrow$ & Action $\uparrow$ & RPE $\downarrow$\\
\midrule
Standard & 245 & 0.36 & 0.50 & 17.48 & 88.4 & 0.75  \\
$+$ MACM  & 228 & 0.36 & \textbf{0.51} & 17.56 & 89.7 & 0.76  \\
Both  & \textbf{179} & \textbf{0.35} & \textbf{0.51} & \textbf{17.72} & \textbf{89.8} & \textbf{0.67 }\\
\bottomrule
\end{tabular}
\label{tab:ablation_main}
\end{table}

\begin{table}[h]
\centering
\begin{minipage}[t]{0.48\textwidth}
\setlength{\tabcolsep}{4pt}
\centering
\caption{\textbf{Ablation on Agent Identity Embedding base frequency.} We compare different base frequencies in Agent Identity Embedding. The result shows that a base frequency of 20 yields a larger rate of decrease that matches the number of agents, leading to better visual and action-following quality.}
\label{tab:aie}
\begin{tabular}{lccc}
\toprule
Config & FVD $\downarrow$ & PSNR $\uparrow$ & Action $\uparrow$ \\
\midrule
base=10k & 234  & 17.53 & 89.2  \\
base=20  & \textbf{228} & \textbf{17.56} & \textbf{89.7} \\
\bottomrule
\end{tabular}
\end{minipage}
\hfill
\begin{minipage}[t]{0.48\textwidth}
\setlength{\tabcolsep}{4pt}
\centering
\caption{\textbf{Ablation on Adaptive Action Weighting.} We evaluate the impact of the  Adaptive Action Weighting mechanism on visual fidelity and action-following performance. The results indicate that Adaptive Action Weighting improves visual quality and action-following ability.}
\label{tab:aaw}
\begin{tabular}{lccc}
\toprule
Config & FVD $\downarrow$ & PSNR $\uparrow$ & Action $\uparrow$ \\
\midrule
w/o AAW & 245 & 17.48 & 88.4  \\
w/ AAW & \textbf{236} & \textbf{17.52} & \textbf{88.6} \\
\bottomrule
\end{tabular}
\end{minipage}
\end{table}

\paragraph{Design of MACM.} MACM consists of Agent Identity Embedding and Adaptive Action Weighting to handle the multi-agent controllability issue. In Tab.~\ref{tab:aie}, we show a suitable parameter selection idea for Agent Identity Embedding in multi-agent control. Moreover, in Tab.~\ref{tab:aaw} we show the necessity of Adaptive Action Weighting. 

Agent Identity Embedding is mainly controlled by the \textit{base frequency} ($\theta$), which determines the frequency decay across agent identity. While the default base frequency of $10000$ is suitable for large language models, it is suboptimal for our multi-agent setting. We conduct ablation experiments over different base values in Tab.~\ref{tab:aie}. With base $=10000$, adjacent agent embeddings are nearly indistinguishable due to minimal angular separation. Reducing the base to $20$ improves performance by matching the number of agents.

The Adaptive Action Weighting mechanism dynamically allocates attention based on the relative importance of each agent's action. As demonstrated in Tab.~\ref{tab:aaw}, Adaptive Action Weighting enhances both action following quality and overall visual quality by prioritizing influential actions over stationary ones.

\paragraph{Design of GSE.} We evaluate four strategies to identify the optimal backbone for maintaining consistency across diverse viewpoints: (1) w/o Global State, which entirely removes environmental context. (2) Wan VAE~\cite{wan2025wan}, which encodes all initial views independently into latent space, introducing minimal gaps with the generation backbone; (3) DINOv2~\cite{oquab2023dinov2}, which utilizes self-supervised features to capture image-level semantic information; and (4) VGGT~\cite{wang2025vggt}, an end-to-end multi-image 3D reconstruction model to capture the relationships between all the observations. Tab.~\ref{tab:ablation_gse} shows that the VGGT backbone yields the best result among other backbones. This highlights the necessity of modeling the shared 3D environment state between multi-view observations.

\begin{table}[t]
\centering
\setlength{\tabcolsep}{7pt}
\caption{\textbf{Ablation on Global State Encoder.} We compare different backbones for Global State Encoder, including native Wan VAE~\cite{wan2025wan}, DINOv2~\cite{oquab2023dinov2}, and VGGT~\cite{wang2025vggt}, against a baseline without a global state encoder. The results show that the VGGT backbone yields the best result among other backbones, improving both visual quality and multi-view consistency.
}
\begin{tabular}{lccccc}
\toprule

\textbf{Global State Encoder} & FVD $\downarrow$ & LPIPS $\downarrow$ & SSIM $\uparrow$ & PSNR $\uparrow$ & RPE $\downarrow$  \\
\midrule
w/o Global State  & 228 & 0.36 & \textbf{0.51} & 17.56 & 0.75 \\
Wan VAE & 256 & 0.36 & 0.50 & 17.38 & 0.71 \\
DINOv2 & 232 & 0.36 & 0.50 & 17.48 & 0.72 \\
VGGT (Ours) & \textbf{179} & \textbf{0.35} & \textbf{0.51} & \textbf{17.72} & \textbf{0.67} \\
\bottomrule
\end{tabular}
\label{tab:ablation_gse}
\end{table}

\begin{figure}[t]
    \centering
    \includegraphics[
        width=\linewidth,
        clip
    ]{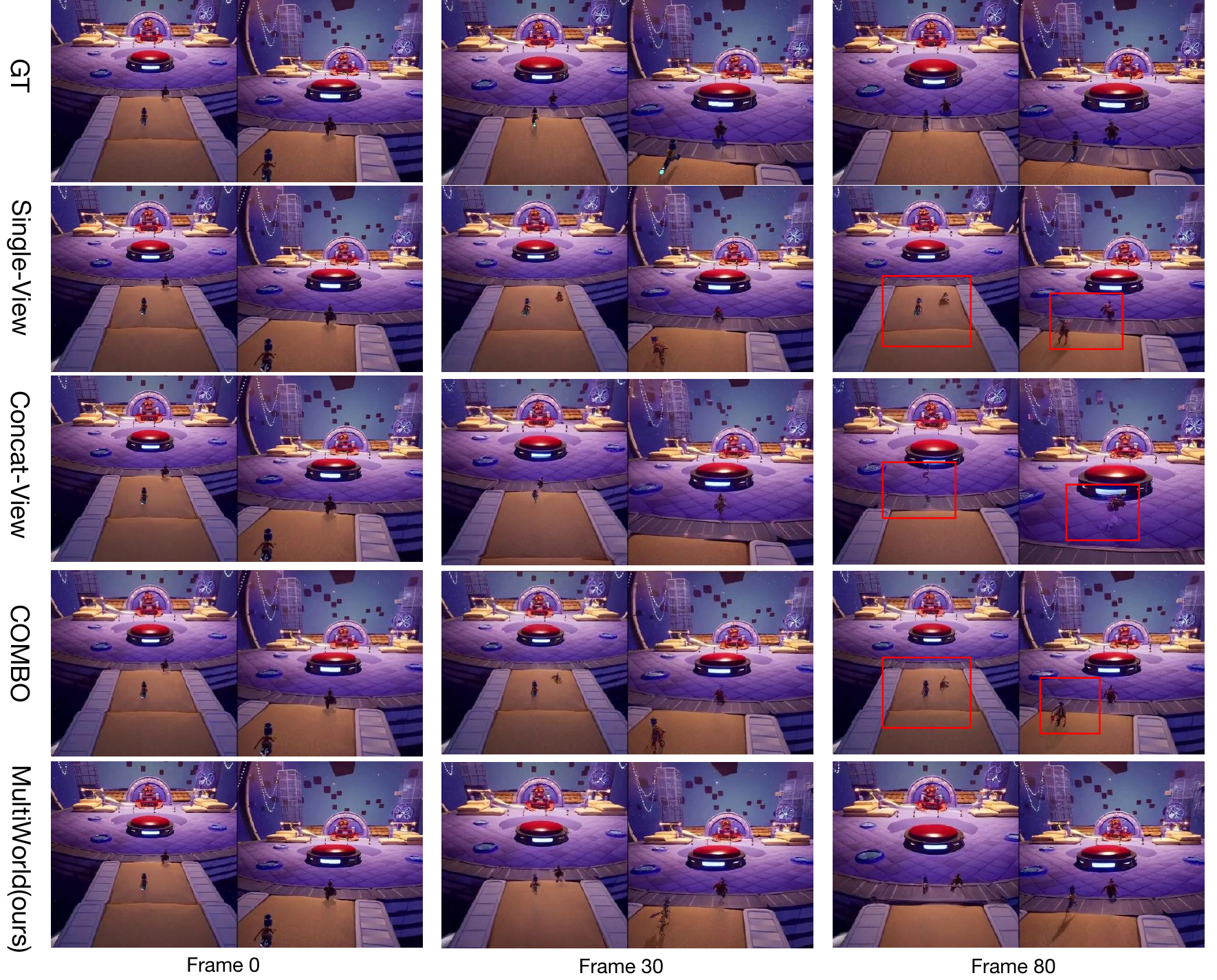}
    \caption{\textbf{Qualitative comparison of multi-agent multi-view video generation in a multi-player video game.} Our method achieves more accurate action following ability and better multi-view consistency compared to the standard baseline, concat-view, and COMBO~\cite{zhang2024combo}. \textcolor{red}{Red boxes} highlight failure cases in competing methods, including inaccurate action execution, agent disappearance, and multi-view inconsistency. MultiWorld can generate videos that faithfully follow given actions with a final state nearly identical to the ground truth.
    } 
    \label{fig:qualitative_comparison}
\end{figure}

\subsection{Qualitative Results}
\label{sec:qualitative}
We provide qualitative comparisons in Fig.~\ref{fig:qualitative_comparison} to illustrate the advantages of our approach. Compared to the standard baseline, concat-view, and COMBO~\cite{zhang2024combo}, MultiWorld has better multi-agent following ability and maintains the environment's structural integrity more consistently. Specifically, competing methods exhibit three typical failure modes: (1) inaccurate action following, where agents fail to execute the specified controls, (2) agent disappearance, where one or more agents vanish during generation, and (3) multi-view inconsistency, where different views have different results. MultiWorld mitigates these issues through the Multi-Agent Condition Module and the Global State Encoder. We also provide additional visualizations of action controllability, agent-environment interactions, and physical consistency in Supp.~\ref{supp:vis}.

\begin{figure}[htbp]
    \centering
    \includegraphics[width=\linewidth, clip]{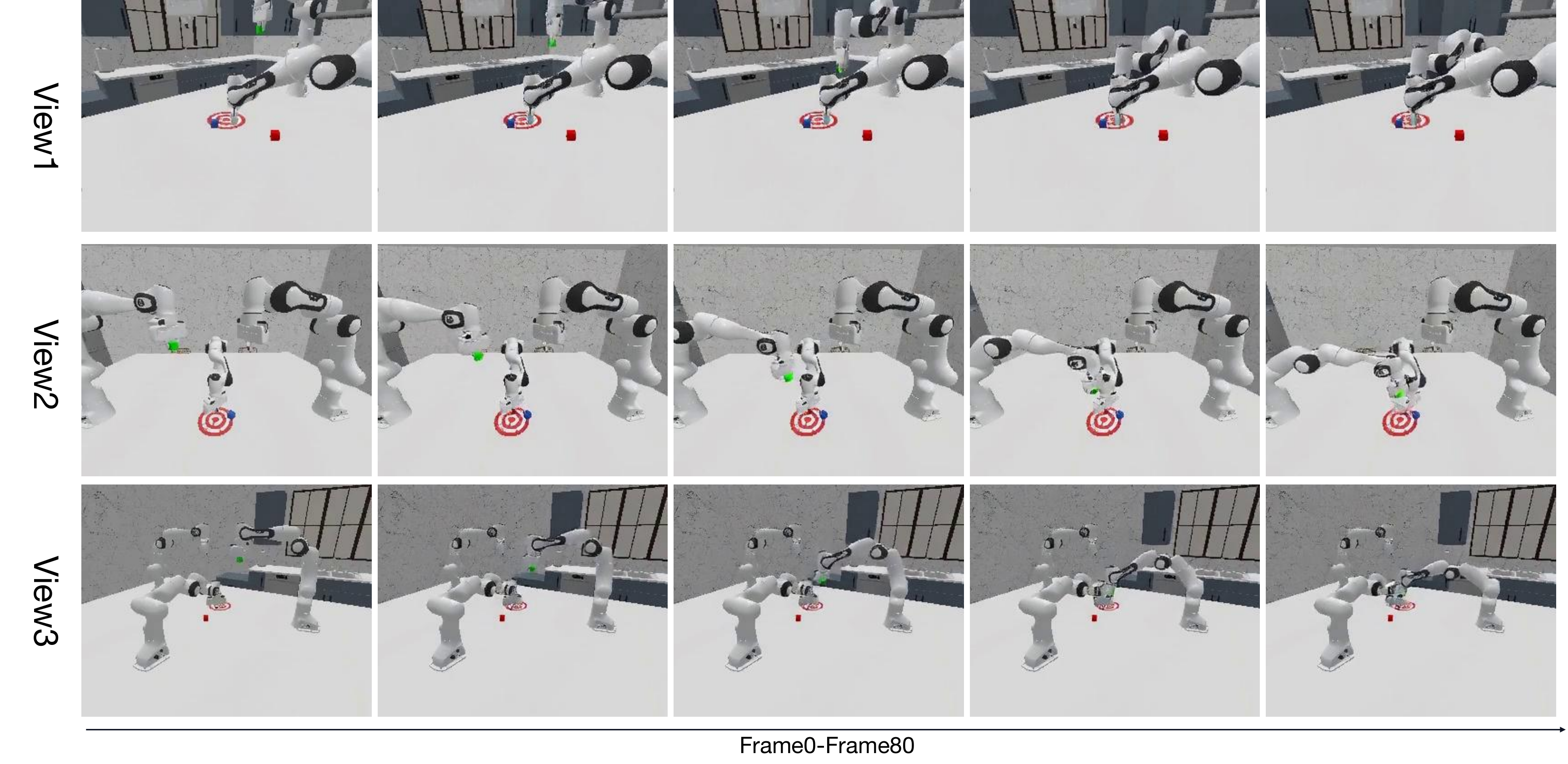}
    \caption{\textbf{Multi-Robot Failure Trajectory Simulation.} MultiWorld simulates realistic cooperative failures, such as inter-robot collisions during collaborative manipulation.}
    \label{fig:supp5}
\end{figure}

\paragraph{Multi-Robot Failure Trajectory Simulation.}
Robotic simulation requires world models capable of simulating failure trajectories. Since the failure trajectories are sometimes difficult to collect, there is a risk of damaging the robots. MultiWorld can perform failure trajectory simulation by enhancing failure trajectory data. Fig.~\ref{fig:supp5} demonstrates MultiWorld's capability to simulate failure trajectories in multi-robot manipulation. Our framework not only generates physically coherent, successful coordination but also accurately captures realistic failure modes, such as inter-robot racing and collisions. The visualization of the success trajectory is provided in Supp.~\ref{supp:vis}.

\begin{figure}[htbp]
    \centering
    \includegraphics[
        width=\linewidth,
        clip
    ]{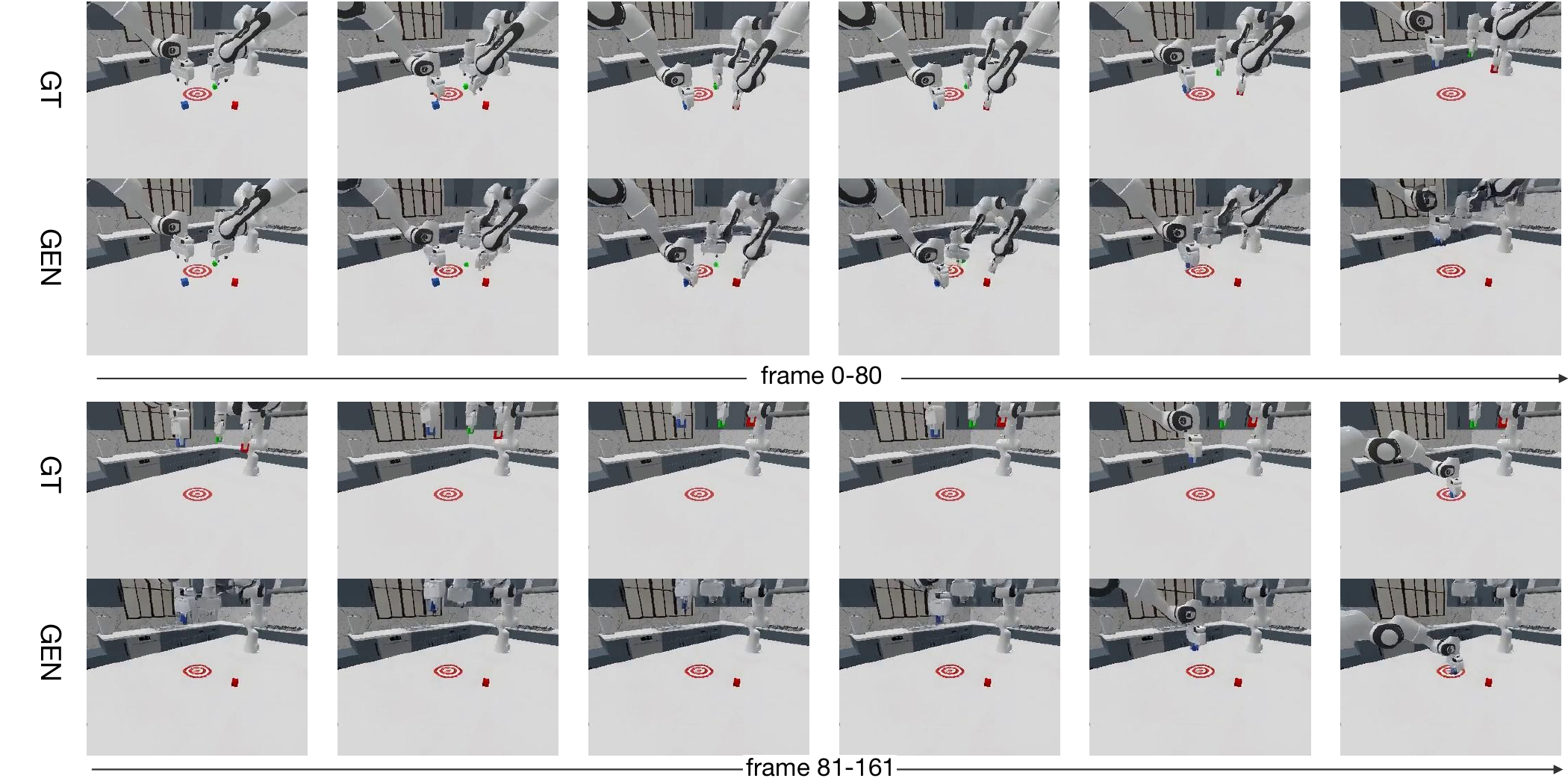}

\caption{\textbf{Long-horizon video generation on multi-robot manipulation task.} Our model autoregressively simulates three robots stacking colored cubes reasonably, maintaining coherence and action accuracy over extended sequences beyond the training context window with little performance drop.}
\label{fig:longvideo}
\end{figure}

\paragraph{Long-Horizon Video Generation.} MultiWorld natively supports autoregressive long-horizon generation as described in Sec.~\ref{sec:inference}. In Fig.~\ref{fig:longvideo}, MultiWorld simulates three robots stacking three little cubes in the correct order fluently without collisions. MultiWorld generates videos up to 2$\times$ longer than the training context window without significant quality degradation, and can extend to 4$\times$ longer with minimal quality loss, enabling long-horizon simulation.

In summary, MultiWorld can simulate a multi-agent, multi-view environment with different properties. It accurately follows action controls and simulates interactions between agents and the environment while maintaining physical consistency during generation and ensuring multi-view consistency.

\section{Conclusion}
This paper presents MultiWorld, a unified framework for multi-agent, multi-view world modeling. MultiWorld is scalable with respect to agent and camera counts, supporting generation on a variable number of agents and camera views. By introducing MACM to ensure multi-agent controllability and GSE to provide environmental information to enhance multi-view consistency. Extensive experiments on multi-player video games and multi-robot manipulation demonstrate the effectiveness of MultiWorld on video quality, action controllability, and multi-view consistency.

\paragraph{Limitation.} While the effectiveness of MultiWorld has been proven, the current scale is still limited. Large-scale training remains unexplored due to computational constraints. 

\paragraph{Future Work.} To advance the applicability of multi-agent multi-view world models to downstream tasks, we plan to investigate real-time multi-agent generation. In addition, multi-agent multi-view video world models impose substantial spatial and temporal memory demands for long-horizon interactions. Future work will therefore explore memory mechanisms for ultra-long multi-agent simulation.

\newpage 

\bibliographystyle{splncs04}
\bibliography{main}

\newpage
\appendix
\setcounter{page}{1}

\section*{\textbf{Appendix for \textit{MultiWorld}: Scalable Multi-Agent Multi-View Video World Models}}
\addcontentsline{toc}{section}{Paper Appendix for \textit{MultiWorld}}

\minitoc 
\section{Dataset Construction}
\label{supp:dataset}

\noindent\textbf{Video Game Dataset.} We collect over 100 hours of human gameplay at 60 FPS with synchronized keyboard and mouse actions. The original resolution of 1440$\times$2560 is downsampled to 320$\times$640 for training.

\noindent\textbf{Robotics Dataset. } We select four multi-robot manipulation tasks: striking, stacking with two robots, stacking with three robots, and passing with four robots. For each task, we collect 1,000 successful and 2,000 failure episodes to reduce success-only bias. Failure episodes are generated by introducing controlled perturbations to successful trajectories. Each episode is recorded from multiple viewpoints at 256$\times$320 resolution.

\noindent\textbf{OpenSource Plan.} We will release a subset of the multi-player video game dataset (It Takes Two) from one chapter to facilitate reproduction. The full game dataset cannot be released due to external constraints. The complete multi-robot cooperation dataset will be made fully publicly available.

\subsection{Dataset Preprocessing for ItTakesTwo}

It Takes Two requires players to complete cooperative tasks using various tools across diverse scenes. During collection, players act naturally, resulting in noisy recordings.

\noindent\textbf{Data Filtering.} We filter frames by: (1) removing non-interactive cutscenes where the game takes control away from players; (2) keeping only side-by-side split-screen frames where one player occupies the left half and the other the right half; (3) removing segments with large camera motion that may cause motion blur or disorientation; (4) removing static action segments where players are idle for extended periods.

\noindent\textbf{Unifying Heterogeneous Action Spaces.} The two players use different controllers (keyboard/mouse vs. gamepad), resulting in heterogeneous action spaces. To avoid separate action encoders, we concatenate the two action types into a single vector. For each player, we mask the other's actions to zero, enabling a single encoder for both agents.

\subsection{Dataset Preprocessing for the Robotics Dataset}

We use the tasks provided in RoboFactory~\cite{qin2025robofactory}.

\noindent\textbf{Failure Case Construction.} To avoid bias caused by including only successful episodes, we construct failure cases based on correct task plans. For example, the strike-cube task includes the following steps: move the arm to a position, open the gripper, close the gripper, move the arm to the target position, and open the gripper again. We preserve the basic operation sequence while introducing controlled randomness at each step to simulate execution errors that can lead to task failure. This strategy ensures that failure episodes remain meaningful in robotic simulation because they are nearly successful rather than purely random. In early experiments, we generated failure cases using completely random actions, which produced meaningless episodes and negatively affected world model training.

\section{Metrics}
\label{supp:metrics}

In this section, we present the detailed implementation of the Reprojection Error (RPE) and the Action Following Ability.

\subsection{Reprojection Error}
\label{supp:RPE}
To evaluate multi-view geometric consistency, we employ Reprojection Error (RPE), a standard metric in visual SLAM. Following the methodology of \cite{duan2025worldscore,wu2025geometryforcing}, we utilize DROID-SLAM \cite{teed2021droid} for scene reconstruction. This process involves extracting frame-to-frame features, then refining camera poses ($G_t$) and pixel-wise depth maps ($d_t$) via differentiable Dense Bundle Adjustment (DBA). By enforcing optical flow constraints, this approach ensures robust structure-from-motion. The RPE is computed as the average Euclidean distance between the observed pixel coordinates and the projected locations of co-visible 3D points. Formally, RPE is expressed as:$$RE= \frac{1}{|\mathcal{V}|} \sum_{(i,j) \in \mathcal{V}} \left\| \mathbf{p}^{*}_{ij} - \Pi(\mathbf{P}_{ij}) \right\|_2$$In this equation, $\mathcal{V}$ represents the set of valid feature correspondences, $\mathbf{p}^{*}_{ij}$ denotes the observed pixel position in the generated frames, and $\mathbf{P}_{ij}$ is the reconstructed 3D point derived from the optimized depth and poses, with $\Pi$ serving as the camera projection function. A lower RPE indicates superior 3D alignment and enhanced spatio-temporal stability, effectively quantifying the geometric integrity of the generated video.

\subsection{Action Following}

\paragraph{Inverse Dynamic Model.}
Following VPT \cite{baker2022vpt}, we train a bidirectional Inverse Dynamics Model (IDM) to infer actions from given video sequences. The IDM architecture uses a ResNet-50 backbone, followed by a temporal processing layer, to predict discrete and continuous actions, respectively. We train the model on the same training dataset for 20 epochs with a learning rate of $10^{-4}$.To evaluate the action-following performance of our multi-agent, multi-view world model, we feed the generated videos into the IDM and compute accuracy for discrete actions and Mean Squared Error (MSE) for continuous actions. As shown in Table~\ref{tab:multi_view_comparison}, we report discrete action accuracy for the Multi-Player Video Game task. For Multi-Robot Manipulation, we report a normalized metric, $100 \times (1 - \text{MSE}(a_{pred}, a_{gt}))$, ensuring that higher values indicate superior action-following consistency.

\section{Implementation Details}

\subsection{Flow Matching}
\label{supp:training}
\paragraph{Training.} We train a neural network $v_\theta$ to predict the velocity field $\mathbf{u}$. The FM objective for view $c$ is
\[
\mathcal{L}_{\mathrm{FM}}
=
\mathbb{E}_{t,\boldsymbol{\epsilon}}
\left[
\left\|
 v_\theta ( \mathbf{x}^\mathbf{t}, \mathbf{t}, \mathbf{a})
- \mathbf{u}
\right\|_2^2
\right],
\]

\paragraph{Sampling.}
At inference time, the sampling follows a simple probability flow ODE:
\[
\text{d}\mathbf{x} = v_\theta ( \mathbf{x}^\mathbf{t}, \mathbf{t}, \mathbf{a}) \cdot \text{d}\mathbf{t}.
\]
In practice, we iteratively apply the standard Euler solver~\cite{euler1845institutionum} to sample video from noise. 

\section{Supplementary Visualizations}
\label{supp:vis}
In this section, we present additional qualitative results that demonstrate MultiWorld's key capabilities.

\begin{figure}[h]
    \centering
    \includegraphics[width=\linewidth, clip]{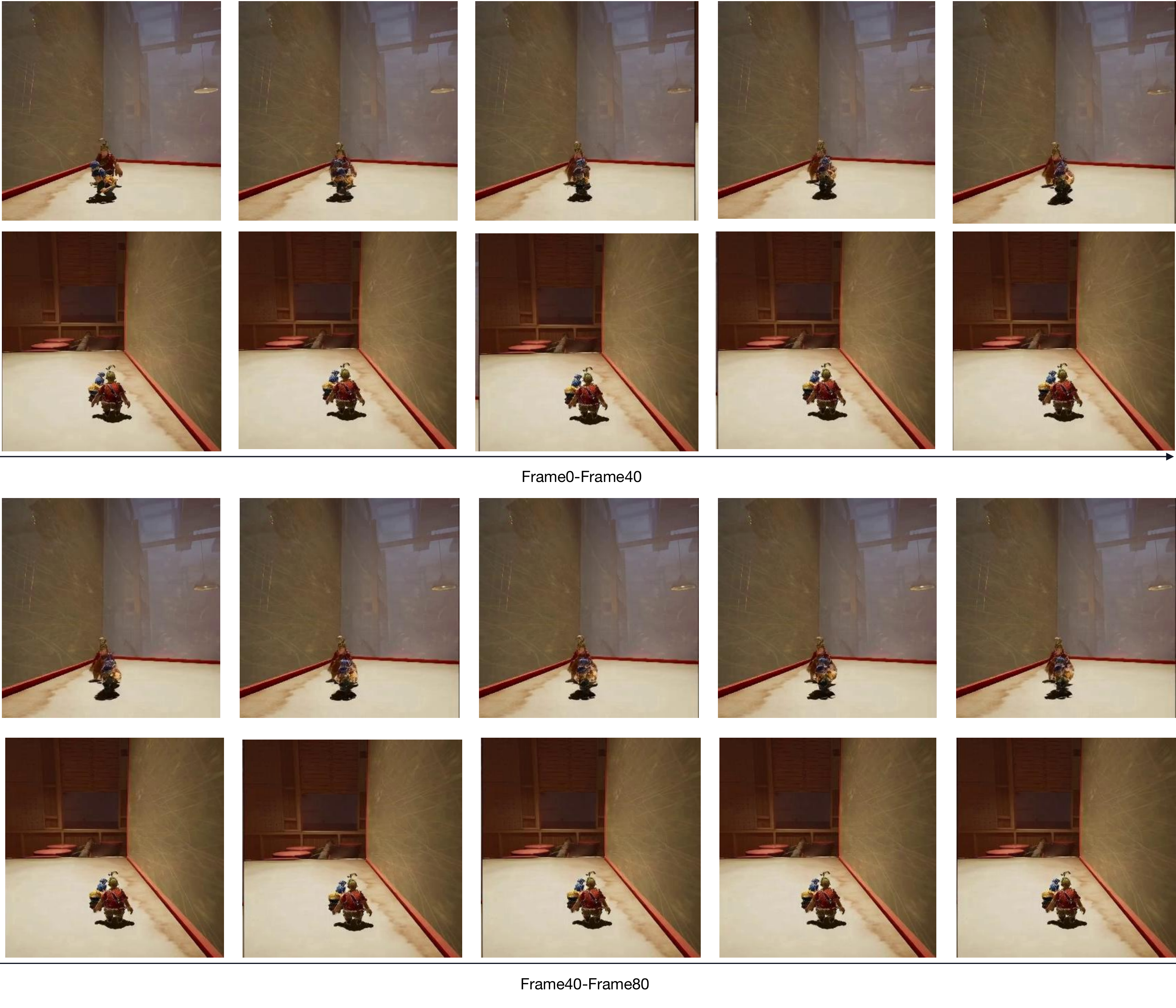}
    \caption{\textbf{MultiWorld Action Controllability.} MultiWorld faithfully generates static videos when given static actions. In contrast, action-conditioned video world models often suffer from action bias, where agents tend to move even without explicit action commands.}
    \label{fig:supp1}
\end{figure}

\paragraph{Action Controllability.} Action-conditioned video models often suffer from action bias, generating unintended motion even with static inputs. As shown in Fig.~\ref{fig:supp1}, MultiWorld faithfully generates static videos in response to zero-action commands, avoiding spurious motion artifacts common in prior approaches.

\begin{figure}[h]
    \centering
    \includegraphics[width=\linewidth, clip]{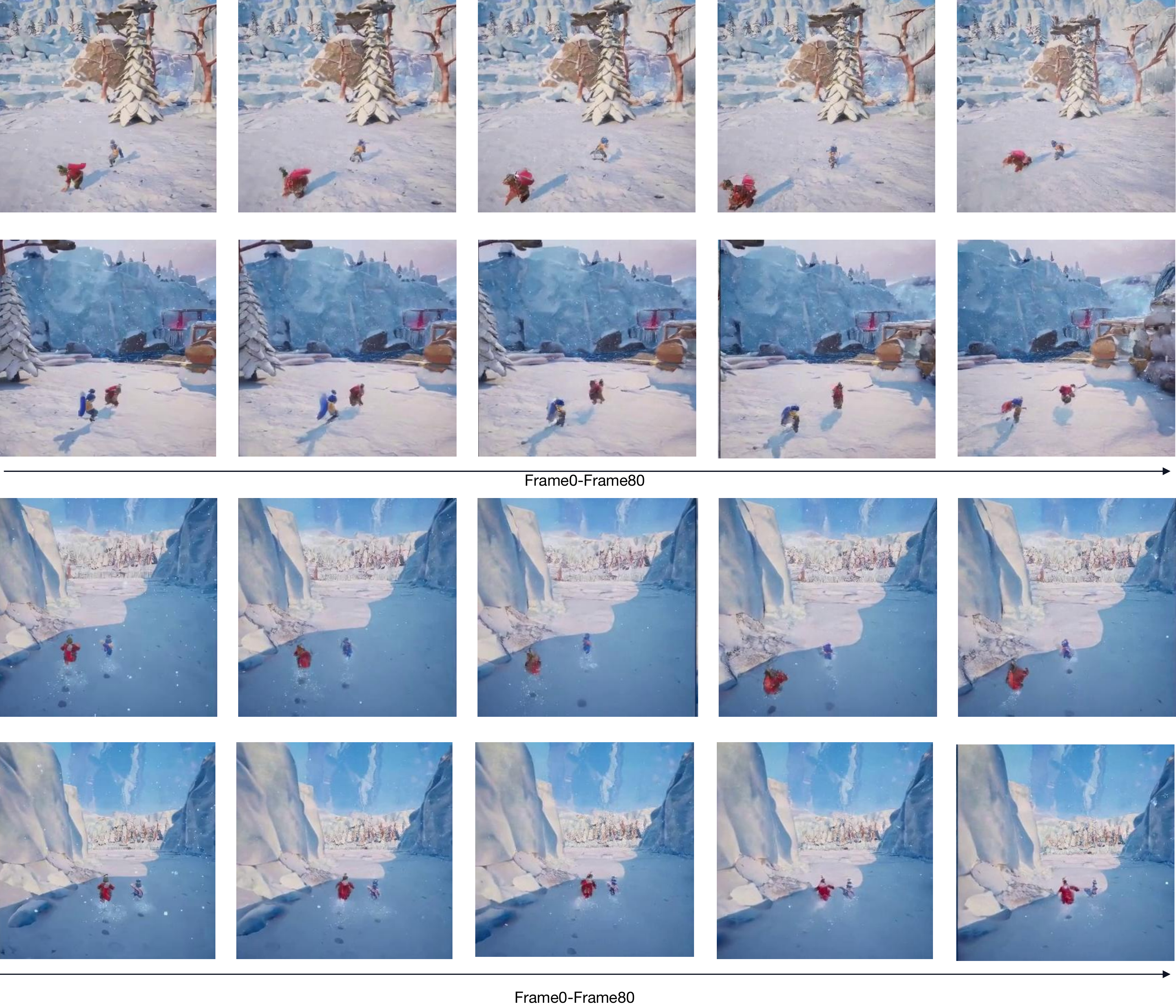}
    \caption{\textbf{MultiWorld Physical Consistency.} MultiWorld generates multi-view videos with consistent physical effects across viewpoints. In the upper case, shadows cast by objects remain consistent in two opposite views. In the lower case, MultiWorld simulates footprints as two agents walk on snow, with the deformation persisting correctly across both camera perspectives.}
    \label{fig:supp2}
\end{figure}

\paragraph{Physical Consistency.} Multi-view consistency requires coherent physical effects across viewpoints. Fig.~\ref{fig:supp2} demonstrates that MultiWorld maintains consistent shadows across opposite views and correctly persists snow footprints in both camera perspectives, indicating it captures underlying physical properties rather than view-specific appearances.

\begin{figure}[htbp]
    \centering
    \includegraphics[width=\linewidth, clip]{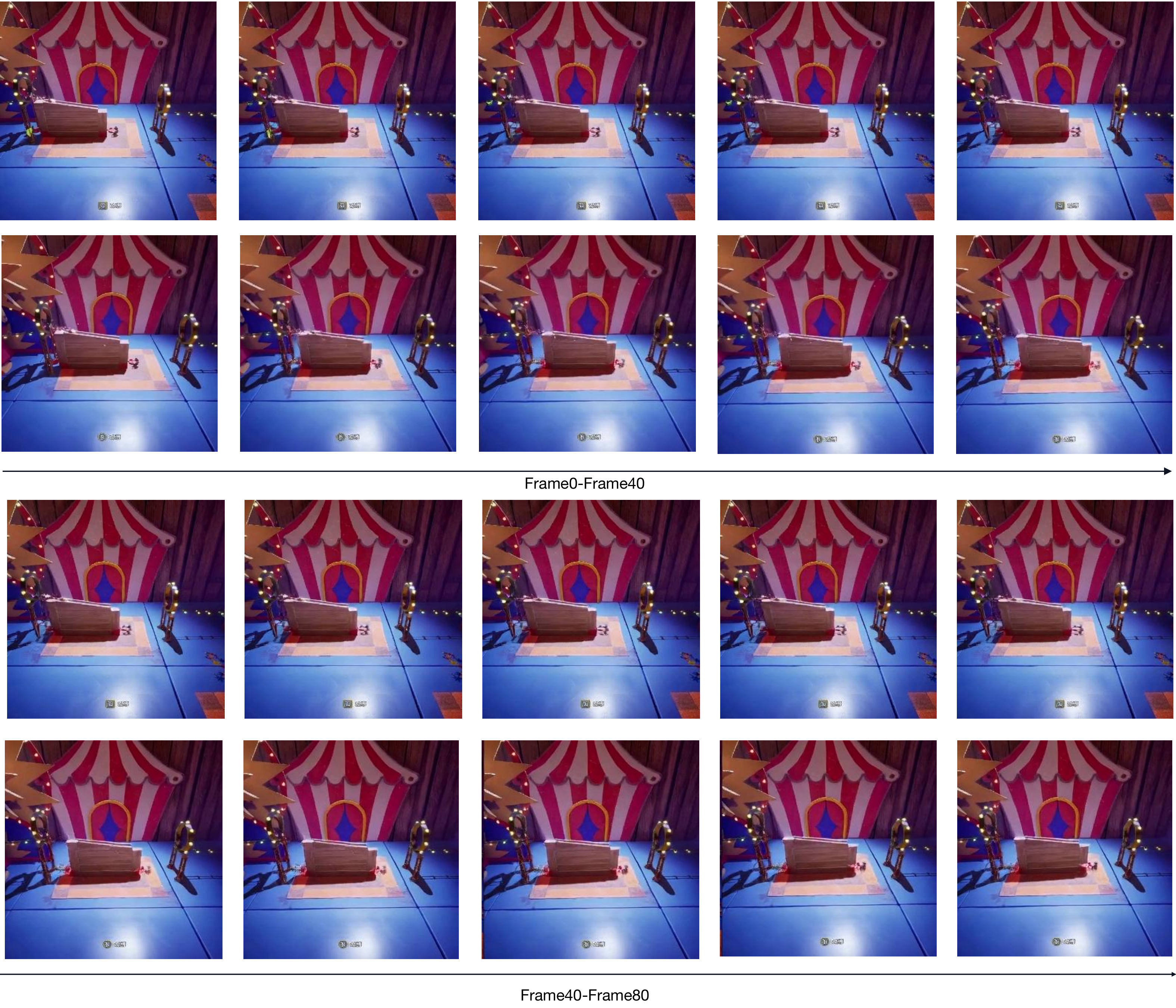}
    \caption{\textbf{MultiWorld Multi-Agent Environment Interactions.} MultiWorld accurately simulates complex environment interactions involving multiple agents. As shown in the figure, one agent pushes while the other pulls a large board from left to right. MultiWorld captures the joint physical influence of both agents on the environment, demonstrating precise action following and consistent environmental dynamics.}
    \label{fig:supp3}
\end{figure}

\paragraph{Multi-Agent Interactions.} Modeling joint environmental influence from multiple agents remains challenging for video generation. Fig.~\ref{fig:supp3} shows MultiWorld simulating coordinated physical interactions in which one agent pushes while another pulls a large board, accurately capturing the combined effects of both agents on the environment.

\begin{figure}[htbp]
    \centering
    \includegraphics[width=\linewidth, clip]{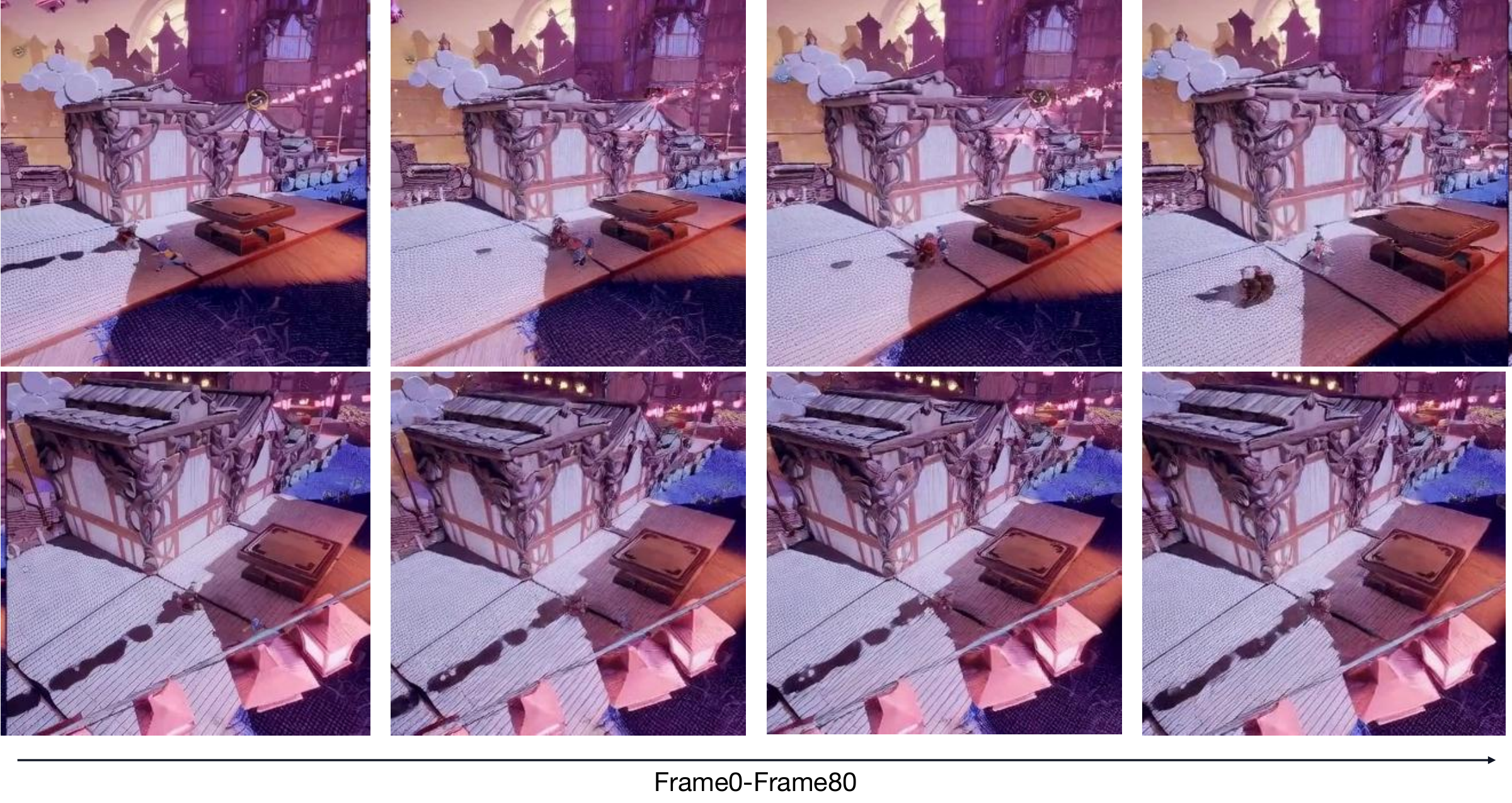}
    \caption{\textbf{MultiWorld Failure Case Analysis.} Agents often appear ambiguous when occupying small regions of the view, due to limited spatial resolution for distant objects.}
    \label{fig:supp4}
\end{figure}

\paragraph{Failure Case Analysis.} Our model also has limitations. Agents often appear in ambiguous shapes when occupying small regions of the view, resulting from limited spatial resolution for distant or small objects, as shown in Fig~\ref{fig:supp4}.

\begin{figure}[htbp]
    \centering
    \includegraphics[width=\linewidth, clip]{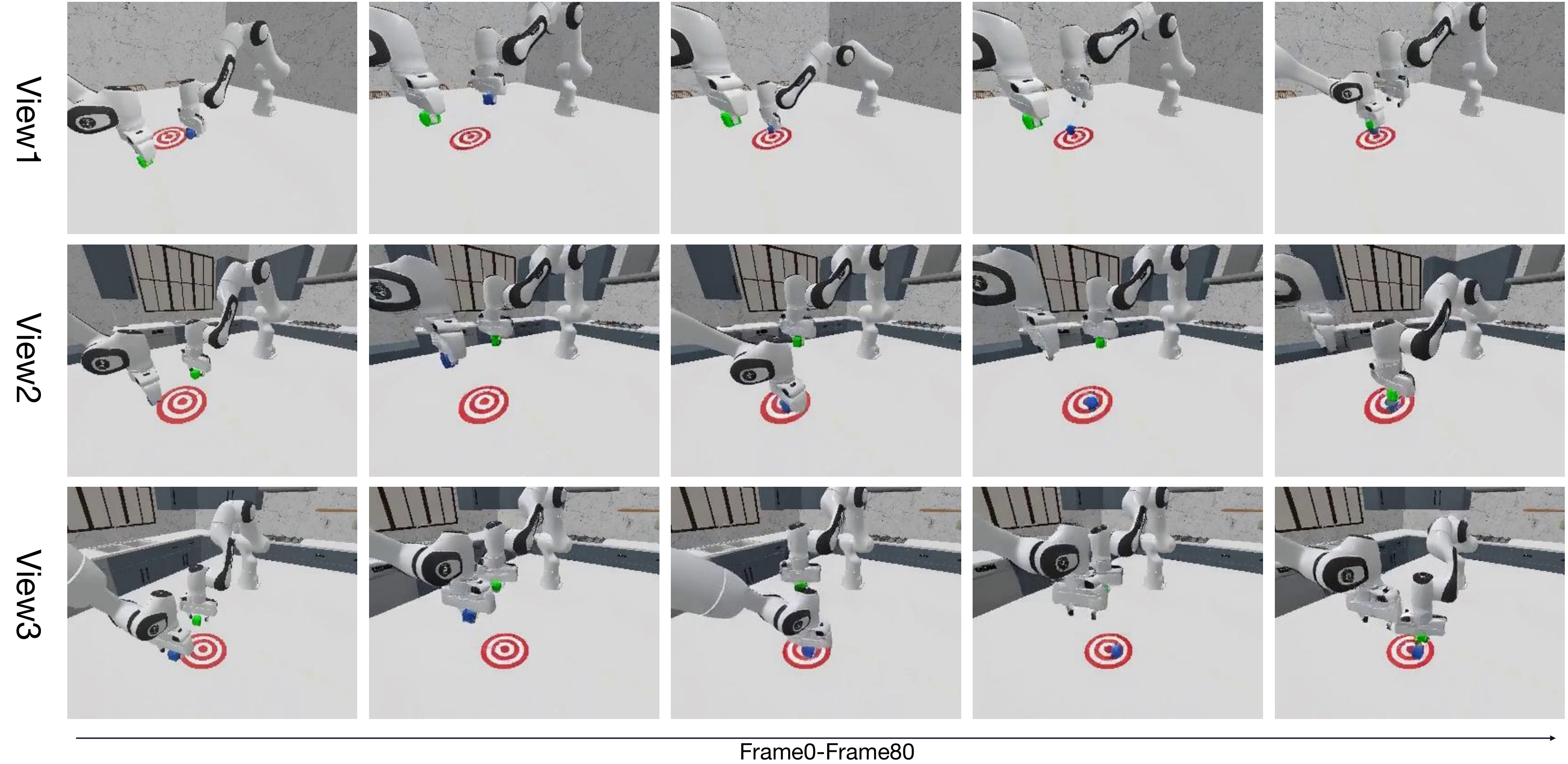}
    \caption{\textbf{Multi-Robot Success Trajectory Simulation.} MultiWorld generates physically plausible cooperative behaviors where robots successfully coordinate to complete manipulation tasks.}
    \label{fig:supp6}
\end{figure}

\end{document}